\documentclass[letterpaper, 10pt, conference]{ieeeconf}  % Comment this line out if you need a4paper
\IEEEoverridecommandlockouts                              % This command is only needed if 
\usepackage{url}
\usepackage{svg}
\usepackage[backend=biber,
            hyperref=true,
            url=false,
            isbn=false,
            doi=false,
            backref=false,
            style=ieee,
            natbib=true,%compatibility aliases
            mincitenames=1,
            maxcitenames=1,
            citestyle=numeric-comp,
            sorting=nyt,%none
            block=none]{biblatex}
            
\usepackage{amsmath,amssymb}
\usepackage{graphicx,color,soul}
\usepackage{adjustbox}
\usepackage{booktabs}
\usepackage{multirow}
\usepackage{verbatim}
\usepackage{multicol}
\usepackage{subfig}
\usepackage{hyperref}
\usepackage{xspace}
\usepackage{textcomp}
\usepackage{gensymb}
\usepackage{marginnote}
\usepackage{siunitx} % Daniel: added based on Jeff's prior papers. (needs to be paired with next line :) -Jeff)
\usepackage{caption}
\graphicspath{{figures/}}
\usepackage[left=0.75in,right=0.75in,top=0.75in,bottom=58pt]{geometry}
\renewcommand\hl[1]{#1} % Uncomment this to turn off highlighting
\newcommand{\ie}{i.e.,\xspace}
\newcommand{\eg}{e.g.,\xspace}

\usepackage{pifont}
\definecolor{britishracinggreen}{rgb}{0.23, 0.53, 0.19} % <-- Dave gets the cool color. :) -Daniel 

\title{\LARGE \bf
Learning to Singulate Layers of Cloth \hl{using} Tactile Feedback 
}

\author{Sashank Tirumala$^{1,*}$, Thomas Weng$^{1,*}$, Daniel Seita$^{1,*}$, Oliver Kroemer$^{1}$,  Zeynep Temel$^{1}$, David Held$^{1}$\\% <-this % stops a space
\thanks{*Equal contribution.}% <-this % stops a space
\thanks{$^{1}$The Robotics Institute at Carnegie Mellon University}%
\thanks{Correspondence: {\tt\footnotesize \{stirumal,tweng,dseita\}@andrew.cmu.edu}}%
}
\begin{document}

\maketitle
\thispagestyle{empty}
\pagestyle{empty}

%%%%%%%%%%%%%%%%%%%%%%%%%%%%%%%%%%%%%%%%%%%%%%%%%%%%%%%%%%%%%%%%%%%%%%%%%%%%%%%%
\begin{abstract}
Robotic manipulation of cloth has applications ranging from fabrics manufacturing to handling blankets and laundry. Cloth manipulation is challenging for robots largely due to their high degrees of freedom, complex dynamics, and severe self-occlusions when in folded or crumpled configurations. 
Prior work on robotic manipulation of cloth relies primarily on vision sensors alone, which may pose challenges for fine-grained manipulation tasks such as grasping a desired number of cloth layers from a stack of cloth. In this paper, we propose to use tactile sensing for cloth manipulation; we attach a tactile sensor (ReSkin) to one of the two fingertips of a Franka robot and train a classifier to determine whether the robot is grasping a specific number of cloth layers. During test-time experiments, the robot uses this classifier as part of its policy to grasp one or two cloth layers using tactile feedback to determine suitable grasping points.
Experimental results over 180 physical trials suggest that the proposed method outperforms baselines that do not use tactile feedback and has better generalization to unseen cloth compared to methods that use image classifiers.
Code, data, and videos are available at \url{https://sites.google.com/view/reskin-cloth}.
\end{abstract}

\section{Introduction}\label{sec:intro}

Cloth manipulation remains an active research area in robotics with significant real world applications, including folding laundry~\cite{maitin2010cloth,unfolding_rf_2014}, assistive dressing~\cite{ra-l_dressing_2018,dressing_2017,deep_dressing_2018,assistive_gym_2020}, bed-making~\cite{seita-bedmaking}, and manufacturing fabrics~\cite{Torgerson1987VisionGR}. Cloth manipulation is challenging because it is difficult to infer the complete configuration of the cloth from robot observations when the cloth is in a crumpled or folded state, due to the high degrees of freedom and self-occlusions~\cite{manip_deformable_survey_2018,grasp_centered_survey_2019}. 
%. Furthermore, modeling the dynamics of cloth (\eg determining the next cloth state given an action) is challenging, particularly with severe self-occlusions~\cite{manip_deformable_survey_2018,grasp_centered_survey_2019}. 

In light of these challenges, researchers have recently proposed numerous data-driven methods for canonical cloth manipulation tasks such as smoothing~\cite{seita_fabrics_2020,lerrel_2020} and folding~\cite{fabricflownet,folding_fabric_fcn_2020,sim2real_deform_2018}. While showing promising results, many prior works focus on top-down grasping of one cloth. Such grasping may be ineffective for manipulation tasks involving multiple cloths, such as picking a desired number of layers of a stack of cloth, because performance is extremely sensitive to the height of the gripper when it grasps. Indeed, a common failure case reported in prior work~\cite{fabricflownet,descriptors_fabrics_2021} is picking the wrong number of layers. Yet, manipulating a specific number of cloth layers is common in daily life, such as in folding and unfolding tasks, or handling piles of stacked clothing in stores. How, then, can robots achieve accurate grasping of multiple layers of cloth?

% Daniel: for all figures, see the following link:
% https://docs.google.com/presentation/d/1K8gHgh9ZgPfHJz2jx81oQ5rLFtZjihxxOj5izai3Ins/edit#slide=id.g1150e038543_0_5
\begin{figure}[t]
\center
\includegraphics[width=0.48\textwidth]{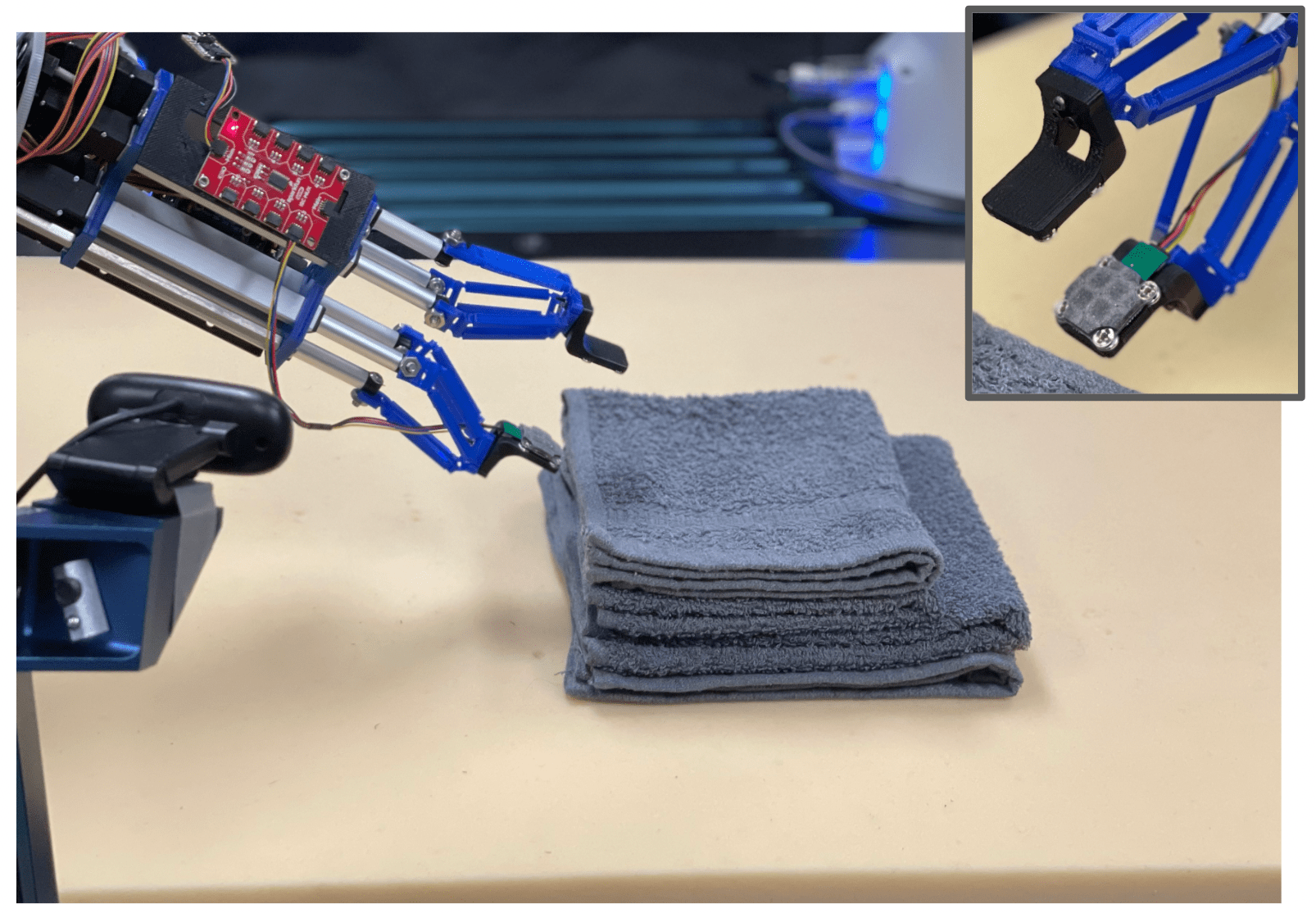}
\caption{
We present a tactile-based cloth manipulation system. The robot utilizes a ReSkin~\cite{bhirangi2021reskin} sensor attached to the lower one of its two fingertips, which is visualized in more detail in the upper right inset. We train a classifier to distinguish among grasping different numbers of cloth layers from tactile feedback (no images are provided as input). The robot then uses this classifier at test time to determine suitable grasping points for obtaining a desired number of cloth layers.
}
\vspace{-5pt}
\label{fig:pull}  % Dave calls it 'pull', Ken calls it 'teaser' :)
\end{figure}

Incorporating tactile sensing is an under-explored direction for deformable object manipulation.
While there has been recent work on optical-based tactile sensors such as GelSight~\cite{yuan_gelsight_2017} and DIGIT~\cite{DIGIT_2020}, these sensors have primarily been applied to cloth \textit{perception}~\cite{vitac_icra_2018,yuan_icra_2018} instead of cloth \textit{manipulation}. Recent work on magnetometer-based sensors such as ReSkin~\cite{bhirangi2021reskin} have benefits over optical sensors, such as lower-dimensional sensor readings, more direct measurements of normal and shear forces, and a compact form factor. However, research into the applications of \textit{magnetometer-based} sensors for deformable object manipulation is currently limited.

In this paper, we study the application of magnetometer-based tactile sensing for deformable cloth manipulation. We focus on precisely grasping and lifting layers of stacked cloth; due to the flexibility of cloth and unpredictable crumpling behavior, this task is challenging while being a well-defined manipulation problem. Furthermore, precise grasping of layers of cloth is a prerequisite for many downstream manipulation tasks (\eg folding cloth in half twice).

We present a robotic system consisting of a 7-DOF Franka arm, a mini-Delta gripper~\cite{DeltaRSS2021}, and a Reskin~\cite{bhirangi2021reskin} sensor on the gripper finger to perform precise cloth grasping (see Fig.~\ref{fig:pull}). The system uses a tactile classifier as feedback for a grasping policy. We show that simple approaches to both classifying tactile data and incorporating feedback into the policy (\eg as a termination condition) work surprisingly well. 

This paper makes the following contributions:  

\begin{enumerate}
    \item A robot hardware system which incorporates ReSkin tactile sensors for cloth manipulation.
    \item A training procedure for developing a classifier based on this hardware to use in a grasping policy.
    \item Experiments showing success on the task of grasping a desired number of cloth layers.
\end{enumerate}

\section{Related Work}\label{sec:rw}

Manipulation of deformable objects such as cloth has a long history in robotics; see~\citet{hang_yin_survey_2021} and~\citet{2021_survey_defs} for representative surveys.

\subsection{Cloth Manipulation Policies}

In early research on cloth manipulation, a common strategy was to utilize a bimanual robot to grip cloth in midair to smooth it using gravity. This standardizes the configuration of cloth and exposes its corners, which can then facilitate planning subsequent manipulation tasks such as smoothing and folding~\cite{maitin2010cloth,unfolding_rf_2014}. Other researchers have relied on using geometric features of cloth, such as by fitting polygon contours to clothing~\cite{laundry2012}. While these works showed impressive results, such approaches may be time-consuming or require strong assumptions on cloth configurations.

With the rise of deep learning, researchers have recently employed data driven techniques to obtain large amounts of interaction data with cloth to learn manipulation policies using powerful function approximators, often with the help of simulators~\cite{corl2020softgym,mujoco}. These works tend to learn quasi-static pick-and-place policies, which allow the cloth to settle between robot actions~\cite{seita_fabrics_2020,seita_bags_2021,fabric_vsf_2020,yan_fabrics_latent_2020,lerrel_2020,fabricflownet,descriptors_fabrics_2021,VCD_cloth,GDOOM,bodies_uncovered_2022}. Other researchers have learned continuous servoing policies~\cite{sim2real_deform_2018}, dynamic policies~\cite{ha2021flingbot} or have explored learning cloth manipulation from purely real world interaction~\cite{folding_fabric_fcn_2020}.

In contrast to these works which employ vision-manipulation policies, we focus on tactile sensing for cloth grasping. 
% These vision-based approaches are orthogonal to our contribution and
% which is orthogonal and can be combined with these vision approaches.%and to theese appwhich can then be used as a starting point for cloth manipulation tasks.

\subsection{Grasping for Cloth Manipulation}

Perhaps the most important part of cloth \emph{manipulation} tasks is cloth \emph{grasping}, since a suitable grasp is necessary for subsequent actions such as dragging or lifting. Defining and identifying ideal cloth grasps remains challenging and is the subject of extensive research~\cite{grasp_centered_survey_2019}.
Early cloth manipulation research focused on vertically smoothing via gravity. A common such grasping strategy to reliably standardize cloth was to hold it with one gripper while iteratively grasping the lowest hanging corner with the other gripper~\cite{kita_2009_icra,kita_2009_iros,maitin2010cloth,cusumano2011bringing}.

Other cloth grasping techniques do not require assuming that the cloth is lifted in midair. For example~\citet{depth_wrinkles_2012} and~\citet{cloth_icra_2015} determine suitable grasping points for cloth on a flat table by detecting wrinkles and edges using depth and classical computer vision techniques. Other applications of cloth manipulation may utilize specialized gripper designs~\cite{three_finger_fabrics_2014} or may simplify the process by assuming that cloth is gripped in advance of the task~\cite{jia_visual_feedback_2018}.

Recently,~\citet{cloth_region_segmentation_2020} study how to robustly grasp cloth using dense segmentation of images to distinguish between edges and interior creases. Their method involves a self-supervised labeling procedure and a sliding grasp. Nonetheless, robustly grasping cloth remains challenging, particularly when the goal is to generalize to a wide variety of types and configurations of cloth. Prior work has reported that a typical failure cases is grasping  the wrong number of cloth layers, particularly when unfolding~\cite{sim2real_deform_2018,seita_fabrics_2020,fabricflownet}. Furthermore, many works employ heuristics
such as hand-tuning the gripper design and grasp depth~\cite{descriptors_fabrics_2021}.
%, such as adding foam to the bottom of a workspace~\cite{fabric_vsf_2020,VCD_cloth} 

%In this work, we utilize tactile sensing to inform suitable grasping points, and apply this to the novel cloth manipulation task of picking a desired number of top layers from a stack of cloth. 
Prior work has also investigated learning to grasp one cloth from a stack using grasping and scooping actions from vision input only~\cite{one_of_stacked_towels_robio2018}, as well as designing a robot system to turn a single book page using vision and force sensing~\cite{deformation_page_turning_2021}. In this work, we consider the novel task of grasping more than one cloth layer, and show the benefits of tactile sensing without requiring vision.

% Daniel: Dave's guide on GitHub suggests to have this figure on page 2, though the intro and related work may push this to page 3 (it seems like that happened with Thomas/Aurora's IROS 2020 paper as well.
\begin{figure*}[t]
\center
\includegraphics[width=1.00\textwidth]{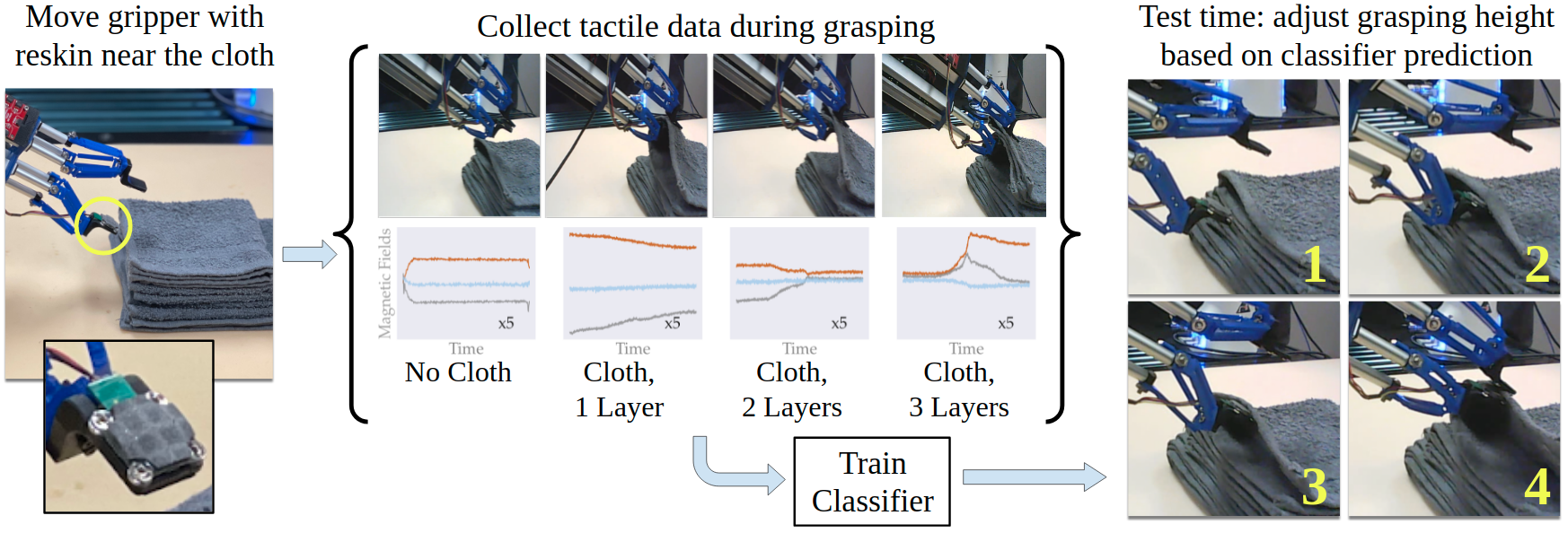}  % finalized? see our slack conversation.
\caption{
The proposed tactile-based cloth manipulation pipeline. A 7-DOF Franka robot uses a mini-Delta~\cite{DeltaRSS2021} gripper with two finger tips, the lower one of which has a ReSkin~\cite{bhirangi2021reskin} sensor (see yellow circle and zoomed-in inset). Using this gripper, we collect tactile data from the ReSkin by performing grasps of different categories: grasping nothing, or pinching 1, 2, or 3 cloth layers (see Fig.~\ref{fig:data} for more examples). 
%Each grasp results in approximately 350 individual 15-D tactile data sensor readings $\mathbf{B}^{(t)}$. 
The graphs above visualize the tactile time series data. At test time, the robot uses the trained tactile-based classifier to grasp a desired number of cloth layers.%, visualized with the 4-frame time-lapse to the right.
%\dave{Note to self - edit this caption}
}
\vspace*{-5pt}
\label{fig:system}
\end{figure*}

\subsection{Tactile Sensor Hardware}

The robotics community has developed numerous tactile sensors. Examples of sensors include the class of optical-based sensors such as GelSight~\cite{yuan_gelsight_2017}, GelSlim~\cite{gelslim_2018}, and DIGIT~\cite{DIGIT_2020}, which have been used for cloth perception. For example,~\citet{yuan_icra_2018} demonstrate how to use active learning to identify where to grasp a garment to classify it among several categories of clothing, and~\citet{yuan_fabrics_cvpr_2017} and~\citet{vitac_icra_2018} study how to combine tactile information with vision to infer properties of cloth. 
Similarly,~\citet{tactile_fabric_learning_2016,tactile_disc_fabrics_ML_2017} use tactile information to classify single fabrics into different material and clothing types using
piezoelectric pressure sensors~\cite{tactile_sensor_for_fabric_2015}.
In addition,~\citet{textile_identification_robio_2017} use tactile sensing and rubbing behavior to classify textiles into one of 18 categories, and can distinguish between 1 or 2 layers.
In contrast to these works, we focus on cloth manipulation instead of pure perception, and additionally focus on fine-grained manipulation which may be challenging with sensors such as the GelSight due to their relatively large size.

% add biotac_2021 if space permits
Other types of robotic tactile sensors include BioTac~\cite{biotac_2019} and stretchable piezoresistive~\cite{rskin_2013} sensors. These sensors are durable, but remain expensive and may not be easily replaceable. 
Research teams have also explored tactile sensing using a customized force-torque sensor~\cite{force_torque_2018} for manipulating deformable blocks~\cite{blind_manip_def_objects_2020}.
To our knowledge, none of these sensors have been used for cloth manipulation tasks.

Recently,~\citet{bhirangi2021reskin} proposed the ReSkin, a class of magnetic sensors which is well suited to machine learning due to its low cost, durability, form factor, ability to cover a large area, and ease of replacement. The researchers demonstrate ReSkin on robotic grasping tasks that involve handling delicate objects such as blueberries and grapes. Due to these advantages and existing applications, we use the ReSkin for novel cloth manipulation tasks that involve fine-grained manipulation of cloth layers.

\section{Problem Statement}\label{sec:PS}

We study the task of grasping a desired number of layers from a stack of cloths. Given a set of at least 3 cloth layers stacked on each other, the goal is to grasp the top $k \in \{1,2\}$ cloth layers. For each \emph{trial} (a given instance of the task), we specify a target value for $k$. We assume a robot has a two-finger gripper where one of the gripper tips is equipped with a \emph{tactile sensor}. We assume each trial begins with the robot's tactile sensor facing a set of edges from a stack of cloth layers, as shown in Fig.~\ref{fig:pull}. A trial is a \emph{success} when the robot grasps exactly $k$ cloth layers and is able to lift its gripper upwards by \SI{4}{\centi\meter} while preserving its grasp of the $k$ layers.

\section{Method}\label{sec:method}

This proposed system for tactile sensing involves designing hardware with tactile data (Sec.~\ref{ssec:hardware}), training a classifier to distinguish grasping cloth layers (Sec.~\ref{ssec:classifier}), then using this classifier for a grasping policy (Sec.~\ref{ssec:grasp-policy}). See Fig.~\ref{fig:system} for the overall pipeline.

\subsection{Hardware}\label{ssec:hardware}

The proposed system uses a ReSkin~\cite{bhirangi2021reskin} sensor, which comprises of a soft magnetized skin and a circuit board with a 5-magnetometer array (see bottom-left inset of Fig.~\ref{fig:system}). The board sits beneath the skin, and any deformations caused by normal/shear forces are read via distortions in magnetic fields. For each of the 5 magnetometers, 3 magnetic flux values $\langle B_X, B_Y, B_Z \rangle$ are reported, corresponding to flux in the X-, Y-, and Z- magnetometer coordinate axes. Concatenating these values for a single time step $t$ results in a 15-dim vector $\mathbf{B}^{(t)} \in \mathbb{R}^{15}$. ReSkin publishes these values at up to \SI{400}{\hertz}.
% The ReSkin sensor~\cite{bhirangi2021reskin} provides five magnetometer values with four numbers each: three magnetic readings $\langle B_X, B_Y, B_Z \rangle$ and one temperature value. 

We attach ReSkin to a finger on a mini-Delta gripper~\cite{DeltaRSS2021}. 
We use the mini-Delta largely due to its length and form factor, since it facilitates grasping a layered stack of cloth folds by approaching it from the side, instead of top-down.
The mini-Delta has 3 DOFs for each finger, and is compliant due to the 3D-printed soft links (blue component in Fig.~\ref{fig:pull}), though in this work, we do not rely on the additional DOFs or on compliance. Our contribution centers on tactile sensing for grasping of cloth layers, and we leave investigation of exploiting additional DOFs and compliance for future work. 
The gripper and attached sensor are mounted on a 7-DOF Franka robot.

\subsection{Grasp Classifier Training}\label{ssec:classifier}

\begin{figure*}[t]
\center
\includegraphics[width=0.95\textwidth]{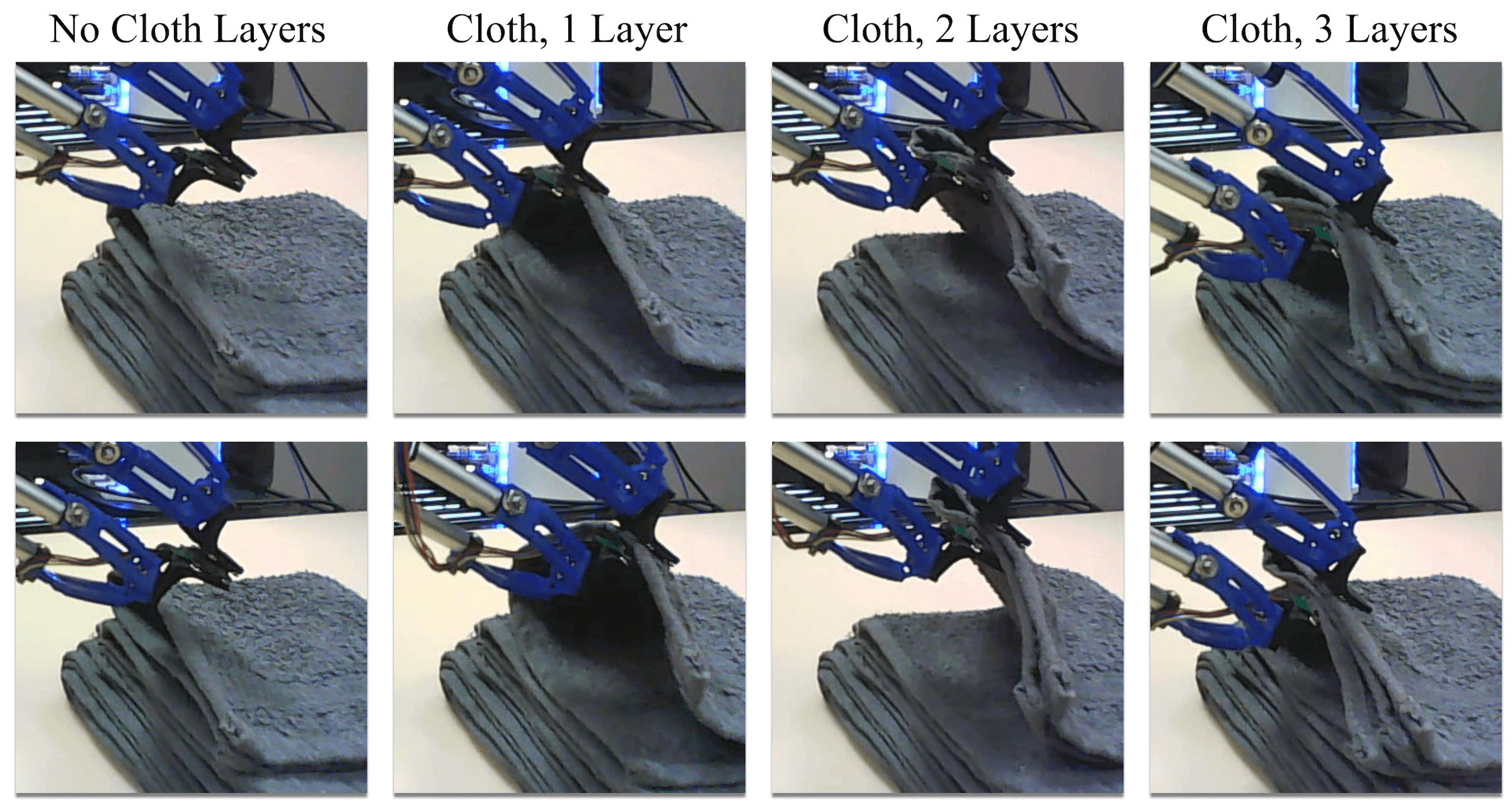}  % 4 classes, smaller size
\caption{
Examples of collecting data for tactile-based classification, with the ReSkin attached to the bottom gripper finger tip. From left to right, we show two examples each of collecting data with (1) contact, but without cloth, (2) 1 cloth layer, (3) 2 cloth layers, and (4) 3 cloth layers. The classifier only takes as input the data collected from the ReSkin sensor $\mathbf{B}^{(t)}$ at any give time step. \hl{The images above are collected with a webcam and are used both to visualize the tactile data collection, and also are the RGB inputs to the image classifiers that we train as baselines for comparison.} See Sec.~\ref{sec:method} \hl{and Sec.}~\ref{sec:experiments} for further details.
}
\vspace*{-5pt}
\label{fig:data}
\end{figure*}

We train a classifier to predict the number of cloth layers grasped to use as part of the grasp policy (Sec.~\ref{ssec:grasp-policy}).
The classifier takes as input a tactile reading from a single time step $\mathbf{B}^{(t)}$. While analyzing sensory data across a time series seems natural for the tactile modality, we find that predictions based on point estimates are surprisingly effective, as we later show in Sec.~\ref{sec:results}.
We do not take proprioceptive data as input, as this modality is not currently available with the mini-Delta gripper: the compliant links can bend from their commanded position given sufficiently high external force, and estimating proprioception for these types of compliant links is an area of active research.

% We do not train a classifier using proprioceptive data (\eg finger tip positions). Such data does not exist with the compliant gripper, which bends when in contact with a sufficiently high external force. This compliance is however advantageous when the arm is trying to slide in between layers of cloth, as the gripper bends and prevents deforming the cloth much unlike a rigid gripper. However, this means a single set of motor angles can correspond to a wide variety of gripper poses.

The classifier uses the tactile readings to predict how the gripper is interacting with the cloth, among 4 classes: (1) pinching with no cloth between the fingers, (2) pinching 1 cloth layer, (3) pinching 2 cloth layers, and (4) pinching 3 cloth layers. 
We limit the number of cloth layers under consideration to 3 to make classification tractable, while also allowing feedback-based policies to recover if they overshoot when grasping two layers. We leave classifying an arbitrary number of layers to future work. %The first class helps to distinguish whether the robot is grasping the cloth at all.

We collect training data in the real world for the classifier due to the lack of a suitable simulator.\footnote{While there has been progress in developing high-fidelity simulators for tactile sensors~\cite{TACTO_2022,Taxim_2022} and for deformables~\cite{corl2020softgym,flex_2014}, simulating both is challenging and to our knowledge has not yet been shown.}
We define a single ``training episode'' as the process of getting a set of tactile data from one grasp. First, a human stacks several layers of cloth with edges facing the gripper. The height at which the robot approaches the cloth is uniformly sampled per attempt within a $\pm \SI{2}{\milli\meter}$ range to collect a variety of grasps.  The robot then approaches the cloth, closes its fingertips to grasp firmly, records ReSkin data during the grasp, then releases. 
Each training episode lasts roughly 5 seconds and produces approximately 350 sensor readings of 15 values each (3 per magnetometer). We visually inspect videos from the recorded data to determine the number of grasped cloth layers, and we label all points from a training episode with the same label, speeding up annotation time and effort. 
See Fig.~\ref{fig:data} for example visualizations of training episodes for all classes.

We then use this collected data to train a classifier to distinguish the numbers of layers grasped from the tactile readings.  We experimented with various types of classifiers, including k-Nearest Neighbor (kNN), SVM, Logistic Regression, and Random Forests, and we found the performance to be fairly similar across classifiers.  For simplicity, we use a k-Nearest Neighbor (kNN) classifier with $k = 10$ neighbors. %(implemented using Scipy~\cite{2020SciPy-NMeth}).

% The ReSkin sensor~\cite{bhirangi2021reskin} provides five magnetometer values with four numbers each: three magnetic readings $\langle B_X, B_Y, B_Z \rangle$ and one temperature value. We omit the temperature so at each time step $t$ within a data collection attempt, we get values from five magnetometers, $\langle B_X, B_Y, B_Z \rangle^{(0,t)}, \ldots, \langle B_X, B_Y, B_Z \rangle^{(4,t)}$, all concatenated to form a single 15-D vector of tactile data $\mathbf{B}^{(t)} \in \mathbb{R}^{15}$ which is then passed to a classifier. 

\subsection{Proposed Grasp Policy}\label{ssec:grasp-policy}

\begin{figure*}[t]
\center
\includegraphics[width=0.95\textwidth]{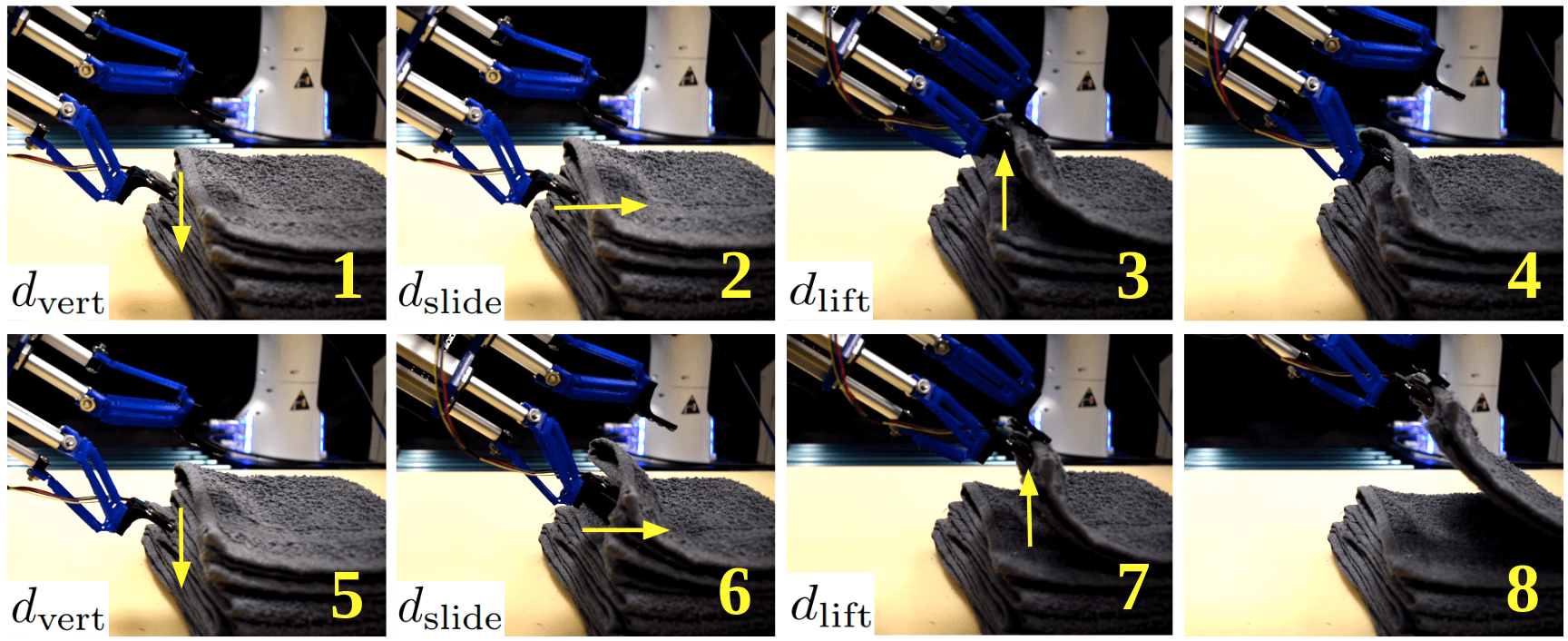} % Daniel: smaller file size
\caption{
The proposed grasp policy parameterization (described in Sec.~\ref{ssec:grasp-policy}), visualized with a frame-by-frame overview of an example trial from the experiments. Each row, consisting of four frames, shows one action. The first part of an action (shown in frames 1 and 5) adjusts the initial gripper height by $d_{\rm vert}$, possibly from prior tactile feedback. The second part of an action (shown in frames 2 and 6) moves towards the cloth stack by some distance $d_{\rm slide}$. Then, the third part (frames 3 and 7) lifts upwards by $d_{\rm lift}$ and closes the grippers. At this point, the robot queries the classifier and may decide to release and re-attempt the grasp (frames 4 and 5) or the robot concludes that it has grasped the correct number of layers and further lifts the cloth to end the trial (frame 8).
}
\vspace*{-5pt}
\label{fig:grasp-param}
\end{figure*}

% Daniel: we are not doing complex grasping.
% For two layers of cloth the parameterization is as follows move down, move horizontally, up, grasp, move down, move horizontally, rub the cloth followed by the final grasp. This reliably generates a good grasp of 2 or more layers of cloth 

Next we describe how we use the above trained classifier to enable the robot to grasp the desired number of layers. We divide the robot trajectory into three parts.
%We propose a straightforward grasp parameterization for each action, which consists of three major parts. 
First, the gripper moves vertically down by a distance $d_{\rm vert}$, then horizontally towards the cloth stack by a distance $d_{\rm slide}$, then lifts up by a distance $d_{\rm lift}$, then closes its gripper tips (see Fig.~\ref{fig:grasp-param} for a visualization). At this point, we record tactile data and classify the number of layers that are grasped.
If the predicted number of grasped layers (according to the classifier) matches the target number of grasped layers, it lifts the gripper further by \SI{4}{\centi\meter} to indicate the end of the trial; otherwise, it resets the gripper back to the starting position and tries again (see below for details). The values of $d_{\rm slide}$ and $d_{\rm lift}$ are tuned and fixed ahead of time by a human operator, while $d_{\rm vert}$ is determined by the policy, as explained below.

The grasping policy uses the output of the  grasp classifier (Sec.~\ref{ssec:classifier}) to determine the vertical distance that the gripper lowers before grasping, $d_{\rm vert}$. %Following Sec.~\ref{sec:PS}, 
For a target number of layers $k$ to grasp, the robot begins at some height with the grippers open, moves towards the cloth stack, and attempts a grasp. If the grasp classifier determines that it has not grasped the correct number of layers, then the robot releases, moves back, and adjusts the gripper height ($d_{\rm vert}$). If the classifier predicts that it has grasped too many layers, $d_{\rm vert}$ is decremented by a small value to decrease the grasp height; if it has grasped too few, $d_{\rm vert}$ is incremented by a small value. The policy continues until either the classifier determines that it has grasped the desired number of layers and ends the trial, or until the maximum number of grasp attempts is reached. 
% We implement k-nearest neighbors (kNN) with 10 neighbors using Scipy~\cite{2020SciPy-NMeth}, and then use the trained model for the grasp policy (Sec.~\ref{ssec:grasp-policy}). 

During each grasp attempt on the physical system, the classifier starts predicting the class once the gripper closes, and stops predicting after the robot lifts by $d_{\rm vert}$. This results in a set of about 160 separate predictions. We use the mode of all the predictions as the final prediction to determine whether to raise or lower the grasp height.

%sent to the controller to determine whether it grasped the correct number of layers or not.

\section{Physical Experiments}\label{sec:experiments}

We evaluate the methods using the physical system described in Sec.~\ref{ssec:hardware}.
% The working tabletop surface is a 1-inch thick foam.
%which has dimension \SI{60}{\centi\meter} by \SI{105}{\centi\meter}. 
The experiments are designed to answer the following questions:

\begin{itemize}
    \item Can magnetometer-based tactile sensing with ReSkin sensors provide sufficient information about grasping a target number of cloth layers?
    \item What are the benefits of the proposed method that uses tactile-based feedback to adjust the gripper height?
    \item Can a classifier trained on tactile feedback generalize to different cloths?
\end{itemize}

\subsection{Experiment Protocol}

\begin{figure}[t]
\center
\includegraphics[width=0.47\textwidth]{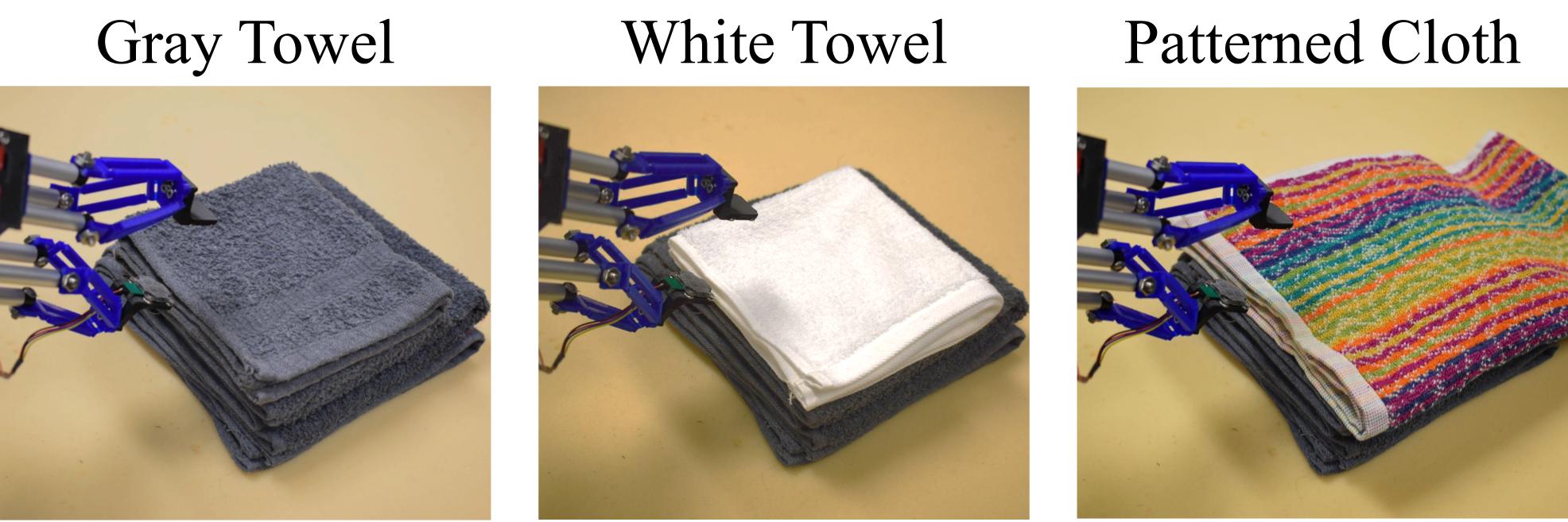}
\caption{
The cloths we use for experiments. We use the gray towel (left) for training, and test on all 3 cloths for evaluation. The white towel and patterned cloth test generalizing to novel cloths. The cloths have thicknesses between 3-5 mm and variation in surface texture and stiffness.
}
\vspace*{-5pt}
\label{fig:cloths}
\end{figure}

We train our tactile classifier on a gray cloth; we then evaluate our system on the gray training cloth and on two other unseen cloths to measure the generalization of our method to new cloths (see Fig.~\ref{fig:cloths}). 
We use the same training data from the gray cloth for all of the tactile-based method variations described in Sec.~\ref{ssec:baselines}.
The tactile data collection results in a total of 18,838 such $\mathbf{B}^{(t)}$ readings.
We train a classifier on \SI{95}{\percent} of the training episodes (to allow for a small validation set). We normalize the tactile data so that each of the 15 features has mean 0 and variance 1 in the training set.

We perform two sets of experiments, in which we set the desired number of cloth layers to grasp as one layer and two layers, respectively.
Each \emph{trial} begins with a human arranging folded cloths on the workspace with edges exposed and facing the robot gripper. 
The number of folded cloths is the same across trials, but variations in the depth of the layers up to \SI{1.5}{\milli\meter} can occur due to slight differences in the initial cloth configuration.
We initialize the robot's gripper at an angle (\SI{30}{\degree}) which increases the likelihood that a horizontal motion can slide the robot finger tips in between layers of cloth. 

Each experiment set consists of comparing several grasping methods (see Sec.~\ref{ssec:baselines}). When running experiments, we randomly sample the method to run in the given trial \emph{after} the cloths have been set, to reduce potential human bias in the data initialization. The robot employs the selected method to grasp the appropriate number of cloth layers. The robot is allowed up to $T=10$ actions per trial, though it can terminate earlier if the classifier estimates that it has grasped the appropriate number of layers. 
%In the open-loop case it terminates after a single trial as it has no access to the classifier. 
Upon termination, the robot lifts the gripper by \SI{4}{\centi\meter} and a human measures this as a success if the correct number of layers are still grasped.
All other cases result in the trial as a failure.

\begin{figure}[t]
\center
\includegraphics[width=0.47\textwidth]{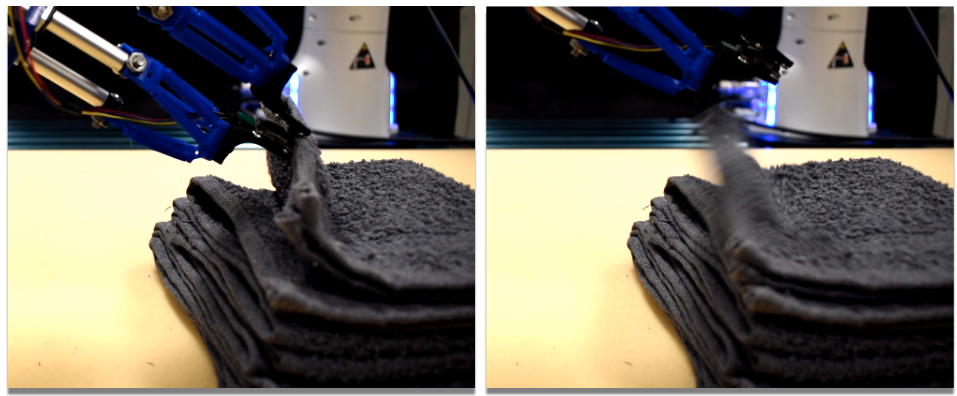}
\caption{
An example grasping failure case of the task. Due to an insufficiently robust grasp when lifting (left), the layers may slip out of the robot's control during the lifting portion (right).
}
\vspace*{0pt}
\label{fig:failure}
\end{figure}

We categorize failures into two types, \emph{prediction} and \emph{grasping} failures. Prediction failures are a result of mis-predictions by the trained classifier, where it either: (1) incorrectly predicts that the robot has grasped the desired number of layers and terminates the trial prematurely, or (2) the classifier incorrectly predicts that the robot has grasped the wrong number of layers, causing unnecessary regrasps and leading to the robot reaching the max number of attempts for the trial. 
Grasping failures are due to either failing to grasp the desired number of layers at the last time step in a trial, or failing to robustly grasp the cloth, such that cloth layers slip out of the robot's control when lifting (see Fig.~\ref{fig:failure}).

\subsection{Methods and Baselines}\label{ssec:baselines}

We evaluate the following methods for grasping 1 and 2 cloth layers:

\begin{enumerate}
\item \textbf{Fixed-Open-Loop}: Initialize the gripper at a fixed height, manually tuned for grasping 1 or 2 cloth layers: $d_{\rm vert}^{\rm (1)}$ and $d_{\rm vert}^{\rm (2)}$ respectively. This method terminates after a single trial as it has no access to feedback. 
\item \textbf{Random-Tactile}: Randomly try different gripper heights within the range $\left[d_{\rm vert}^{\rm (2)} - \text{\SI{2}{\milli\meter}}, d_{\rm vert}^{\rm (1)} + \text{\SI{2}{\milli\meter}}\right]$ until the tactile classifier determines that the correct number of layers have been grasped.
\item \textbf{Random-Image}: Same as Random-Tactile, but uses an image classifier (instead of a tactile classifier) to determine when the correct number of layers has been grasped. The image classifier is an 18-layer ResNet~\cite{resnets_2016} pre-trained on ImageNet and finetuned on the images \hl{(see Figure}~\ref{fig:data}) from the same training episodes used to train the tactile classifier.
\item \textbf{Feedback-Image}: Same as Feedback-Tactile (our method, below) except with the image classifier.
\item \textbf{(Ours) Feedback-Tactile}: Initialize the gripper height to $d_{\rm vert}^{\rm (1)} + \text{\SI{2}{\milli\meter}}$;
%, the maximum of the height range of the Random- baselines. 
use the grasp policy described in Sec.~\ref{ssec:grasp-policy} to adjust the height per grasp ($\pm$\text{\SI{2}{\milli\meter}}) based on the tactile classifier predictions.
\end{enumerate}

%\dave{The below information was said earlier so removing it here to save space}
% We also test with new cloths to measure unseen cloth generalization for both 1-layer and 2-layer grasping. We evaluate the two feedback-based policies, Feedback-Tactile and Feedback-Image, using the same tactile and image classifiers trained on the gray cloth. 

\section{Results}\label{sec:results}

We first present results from training a classifier on ReSkin data followed by physical experiment results in which we run 10 trials for each method and condition.

\subsection{The Tactile Classifier}

\begin{table}[t]
  \setlength\tabcolsep{5.0pt}
  %\begin{adjustbox}{width=\columnwidth,center}
  \centering
  \footnotesize
  \begin{tabular}{@{}lrrrr@{}}
  \toprule
  Class \textbackslash \; Prediction & \multicolumn{1}{c}{0}  & \multicolumn{1}{c}{1} & \multicolumn{1}{c}{2} & \multicolumn{1}{c}{3} \\
  \midrule
  0 (0 Layers) & 1.000 & 0.000 & 0.000 & 0.000 \\
  1 (1 Layer)   & 0.000 & 0.999 & 0.000 & 0.001 \\
  2 (2 Layers)  & 0.030 & 0.003 & 0.866 & 0.100 \\
  3 (3 Layers)  & 0.128 & 0.256 & 0.138 & 0.478 \\
  \toprule
  \end{tabular}
  %\end{adjustbox}
  \caption{
  The average normalized confusion matrix from the cross-validation training results for the k-nearest neighbor classifier we use for tactile-based experiments.
  }
  \vspace*{-5pt}
  \label{tab:classifier}
\end{table}

%\dave{This paragraph should not be in the results section - they should be moved either to methods or to physical experiments}
% As discussed in Sec.~\ref{ssec:classifier}, we train a classifier to distinguish between four classes based on ReSkin tactile data from a given time step $\mathbf{B}^{(t)} \in \mathbb{R}^{15}$ as input. 

To better understand the kNN performance, we perform 100 folds of cross-validation and average the validation performance. Each entire training episode is assigned to either the training or validation set.
% Since we collect data with a sequence of separate ``training episodes'', and we train classifiers on data from single time steps $\mathbf{B}^{(t)}$ within these, each fold has a randomized assignment of ``training'' versus ``validation'' episodes, sampled with an 80-20  split. This can mitigate the effect of correlated data, as all the $\mathbf{B}^{(t)}$ in one episode must be entirely assigned to training or entirely assigned to validation.

Table~\ref{tab:classifier} demonstrates the average normalized confusion matrix obtained from these 100 cross-validation runs, and also reports the average per-class accuracy. We also computed the average balanced accuracy metric~\cite{2020SciPy-NMeth} to consider the data imbalance and obtain $\mathbf{0.84\pm0.06}$. 
%\textbf{0.8357 +/- 0.06}. 
Inspecting the confusion matrix, we find that the tactile classifier can classify classes 0 (\ie pinching with no cloth between the fingers) and 1 (\ie pinching 1 cloth layer) with extremely high effectiveness.
%; the kNN perfectly classifies all the ground-truth class 0 data, and is near-perfect (0.999) for ground-truth class 1 data. 
Results for classes 2 and 3 suggest that identifying 2 and 3 cloth layers is more challenging.
%, motivating the choice to only consider up to 3 layers in this work.

%\dave{The below paragraphs should not be in the results section - they should be moved either to methods or to physical experiments}

%For experiments on the physical system, we train a classifier on \SI{95}{\percent} of the training episodes to use more of the data. During each grasp attempt during the \daniel{Thomas: did you mean to complete this?}

%On the physical system, the classifier predicts predicts the class for around 0.4 seconds which corresponds to around 160 steps of ReSkin data. The prediction starts once the gripper closes and ends before the gripper starts it's next action. The mode of all the predictions is the final prediction sent to our controller to determine whether it grasped the correct number of layers or not.

\subsection{Grasping 1 Cloth Layer}\label{ssec:grasp-01}

% Experiment A: clean arrangement of cloths 
% Hypothesis: clean arrangement would help the manual height setting work well
% Variants:
% manual-no noise
% manual-noise (secondary)
% feedback-no noise
% feedback-noise (secondary)
% feedback-finetuned (with or without noise)
% random
% This is basically the setup for our current table II and III experiments
% NOte: change feedback controller to sample a random range between max height and 1 cloth instead of just using the max height

% Experiment B: slightly rumpled arrangement of cloths 
% rumpling: pressing the cloths down a bit more, fanning them out a bit more, basically adding more height variation to the layers
% Hypothesis: Rumpling should reduce performance of the manual height method
% Variants:
% manual-no noise
% manual-noise (secondary)
% feedback-no noise
% feedback-noise (secondary)
% feedback-finetuned (with or without noise)
% random (image and tactile)
% These experiments would probably supersede the above experiment with clean arrangement of cloths, we would have to describe our rumpling procedure in the experimental setup, and we may want to include the original experiments in the appendix to show how much rumpling affects the results. Since we are sampling which method to run we will not be biasing any particular method with rumpling

% experiment plan
% collect finetuning data
% Run table II IC first (non rumpled)
% Run Exp B with rumpled cloth: (manual-no noise, feedback no-noise, random-image, random-tactile) on 1 and 2 layers
% Run noisy versions of manual and feedback
% Run feedback-finetuned
\begin{table*}[t]
  \setlength\tabcolsep{6.0pt}
  %\begin{adjustbox}{width=\columnwidth,center}
  \centering
  \footnotesize
    \begin{tabular}{llrrrc}
    \toprule
        \multirow{2}{*}{Cloth Type} & 
        \multirow{2}{*}{Method} &  
        Success  &  
        Prediction &  
        Grasp  & 
        \multirow{2}{*}{Attempts $\downarrow$} \\
        % \multirow{2}{*}{Attempts} \\
        {} & 
        {} & 
        Rate $\uparrow$ & 
        Failure & %$\downarrow$ & 
        Failure & \\ %$\downarrow$ & \\
    \midrule
    % \cmidrule(lr){1-2}\cmidrule(lr){3-3}\cmidrule(lr){4-6}
    \multirow{5}{*}{Gray Towel (Train)} 
        & Fixed-Open-Loop        &                6/10 &                               - &                         4/10 &           1 (fixed)\\
        & Random-Image     &                5/10 &                              5/10 &                          0/10 &           1.8$\pm$0.7 \\
        & Random-Tactile   &                6/10 &                              3/10 &                         1/10 &           4.8$\pm$2.8 \\
        & Feedback-Image   &                8/10 &                              2/10 &                          0/10 &           2.3$\pm$0.8 \\
        & Feedback-Tactile &               \textbf{10/10} &                               0/10 &                          0/10 &           3.1$\pm$1.0 \\
    \midrule
    % \cmidrule(lr){1-2}\cmidrule(lr){3-3}\cmidrule(lr){4-6}
    \multirow{2}{*}{White Towel (Generalization)} 
        &  Feedback-Image   & 3/10 & 5/10 & 2/10 & 1.6$\pm$0.5 \\
        & Feedback-Tactile & \textbf{8/10} & 0/10 & 2/10 & 2.3$\pm$0.8 \\
    \midrule
    % \cmidrule(lr){1-2}\cmidrule(lr){3-3}\cmidrule(lr){4-6}
    \multirow{2}{*}{Patterned Cloth (Generalization)} 
        & Feedback-Image   & 2/10 & 8/10 & 0/10 & 5.1$\pm$4.3 \\
        & Feedback-Tactile & \textbf{7/10} & 2/10 & 1/10 & 4.6$\pm$3.2 \\
    \bottomrule
    \end{tabular}
  %\end{adjustbox}
  \caption{
  Results for the first set of physical experiments described in Sec.~\ref{ssec:grasp-01} with grasping at 1 cloth layer. We run all methods for 10 trials each and report the success rate, the failure types (grasping and prediction), and the average number of grasp attempts per trial. 
  %The results suggest that policies using tactile-based feedback to determine gripper height outperform competing methods, and that tactile-based classifiers are more helpful than image-based classifiers for in-domain and out-of-domain cloth.
  }
  \vspace*{-5pt}
  \label{tab:grasping-one}
\end{table*}  

In the first set of physical experiments, we report the success and failures of methods on grasping and lifting the top layer of cloth from a stack. 
See Table~\ref{tab:grasping-one} for results. Our method, Feedback-Tactile, succeeds at grasping one layer of cloth in all 10 trials, whereas all competing ablations have lower success rates. Methods with the tactile classifier outperform those using the image classifier, with most failures attributed to mis-prediction rather than poor grasping.

%The choice of policy also affects performance: 
The fixed-height open loop method (Fixed-Open-Loop) poorly handles variations in the initial cloth configuration. There can be up to \SI{1.5}{\milli\meter} variation in the height of the cloth stack based on how they are placed at the start of the trial, which can lead to failures in the open loop grasping method.
%, possibly because the deformable nature of cloth means it takes on slightly different configurations per trial, so the same grasp action can have varying effects on the cloth.
Both random grasping approaches, Random-Image (5/10) and Random-Tactile (6/10) have lower success rates compared to using feedback-based height adjustment with Feedback-Image (8/10) and Feedback-Tactile (10/10).

For testing generalization, Feedback-Tactile significantly outperforms Feedback-Image on the white towel and patterned cloth. Feedback-Tactile obtains 8/10 and 7/10 success rates for the white towel and patterned cloth, respectively, while Feedback-Image only succeeds in 3/10 and 2/10 trials.

We have analyzed the failure types of each method in Table~\ref{tab:grasping-one}.
Grasping failures are rare for most methods on 1-layer grasping; grasping failures can occur if the robot does not robustly grip the cloth, and cloth slips out of the grasp when the robot lifts it (see Fig.~\ref{fig:failure}).
Our method (Feedback-Tactile), also has few prediction failures when generalizing to unseen cloths compared to Feedback-Image.

% When generalizing to unseen cloths, the number of prediction failures for Feedback-Tactile is low, and there are much fewer prediction failures compared to Feedback-Image. 
% \dave{What conclusions can we draw here?}

%Prediction failures occur either when the classifier predicts a false positive and terminates the trial prematurely, or false negatives result in the system reaching the max number of attempts for the trial.

\subsection{Grasping 2 Cloth Layers}\label{ssec:grasp-02}

% Data directory 
\begin{table*}[t]
  \setlength\tabcolsep{5.0pt}
  %\begin{adjustbox}{width=\columnwidth,center}
  \centering
  \footnotesize
    \begin{tabular}{llrrrc}
    \toprule
        \multirow{2}{*}{Cloth Type} & 
        \multirow{2}{*}{Method} &  
        Success &  
        Prediction  &  
        Grasp  & 
        \multirow{2}{*}{Attempts $\downarrow$} \\
        {} & 
        {} & 
        Rate $\uparrow$ & 
        Failure & %$\downarrow$ & 
        Failure & \\ %$\downarrow$ & \\
    \midrule
    \multirow{5}{*}{Gray Towel (Train)} 
        & Fixed-Open-Loop        &                7/10 &                               - &                         3/10 &           1 (fixed) \\
        & Random-Image     &                6/10 &                              1/10 &                         3/10 &           5.3$\pm$3.0 \\
        & Random-Tactile   &                4/10 &                              4/10 &                         2/10 &           6.0$\pm$3.0 \\
        & Feedback-Image   &                \textbf{9/10} &                               0/10 &                         1/10 &           4.7$\pm$0.9 \\
        & Feedback-Tactile &                7/10 &                              1/10 &                         2/10 &           6.4$\pm$2.6 \\
    \midrule
    \multirow{2}{*}{White Towel (Generalization)}
        & Feedback-Image   &                 0/10 &                              8/10 &                         2/10  &        9.2$\pm$1.8 \\
        & Feedback-Tactile &                \textbf{4/10} &                              2/10 &                         4/10 &            5.0$\pm$3.4 \\
    \midrule
    \multirow{2}{*}{Patterned Cloth (Generalization)}
        & Feedback-Image   &                 0/10 &                             10/10 &                          0/10 &          10.0$\pm$0.0 \\
        & Feedback-Tactile &                \textbf{1/10} & 3/10 &                         6/10 & 6.4$\pm$3.6 \\
\bottomrule
\end{tabular}
  %\end{adjustbox}
  \caption{
  Experimental results for grasping at the top 2 cloth layers as described in Sec.~\ref{ssec:grasp-02}. Besides the change of 1 to 2 layers, the results are formatted in the same way as in Table~\ref{tab:grasping-one}. 
  %Results suggest that the image classifier (Feedback-Image) outperforms tactile-based methods on the gray towel, possibly because tactile sensors have more mis-predictions due to difficulty with classifying two layers. The tactile-based methods, however, exhibit stronger generalization results on the white towel and patterned cloth. \thomas{Conclusions for 2 layer: Image-based classifier surprisingly performs better than tactile on gray towel, because mispredictions are more common for tactile; it's harder for tactile to classify two layers than one. The image classifier performs worse than tactile on generalization, though both perform poorly. Since we have space we can leave these tables as double column width? Explain bolding and why we bold success rate but not the failures that are broken out}
  }
  \vspace*{-10pt}
  \label{tab:grasping-two}
\end{table*}  

In the next set of experiments, we evaluate grasping and lifting the top two layers of cloth. 
The results in Table~\ref{tab:grasping-two} suggest that the methods achieve success rates similar to 1-layer grasping (Table~\ref{tab:grasping-one}) for the gray towel, but performance is lower on the unseen cloths.
While Feedback-Image performs slightly better than Feedback-Tactile on the gray towel, 
Feedback-Tactile performs slightly better on unseen cloths.
%, but many failures can be attributed to grasping rather than prediction error. 
% However, the image based classifier is unable to generalize to the unseen towels, whereas the tactile classifier can generalize better. 

% See Fig.~\ref{fig:grasp-param} for a representative frame-by-frame overview of the proposed Feedback-Tactile policy, where it is able to use the classifier to adjust its actions from initially grasping one layer to ultimately grasping two layers, as desired. More examples and videos are available on the project website.

\begin{figure}[t]
\center
\includegraphics[width=0.480\textwidth]{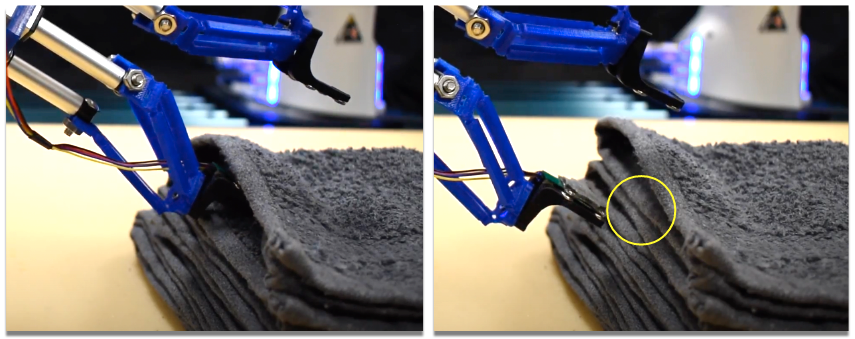}
\caption{
A qualitative example of how the task is challenging, particularly with grasping two layers. Because of the horizontal motion of the gripper, layers of cloth can be pushed apart (left), creating air pockets between the top and second layer after the action has finished (right). This gap makes it easier to grasp the top layer but harder to grasp the top \emph{two} layers, due to a smaller gap between the second and third layers (see overlaid yellow circle).%, and this increases the likelihood that the trial will ultimately end in a failure.
}
\vspace*{-5pt}
\label{fig:task-hard}
\end{figure}

Table~\ref{tab:grasping-two} shows that both prediction and grasping failures lead to errors for our method (Feedback-Tactile), though grasping failures are more common (accounting for 2/3 of our total failures).
The higher incidence of grasp failures by our method in this experiment suggests that 2-layer grasping is more difficult than 1-layer grasping. Fig.~\ref{fig:task-hard} highlights some challenges with grasping two layers; for example, we observe that failures tend to occur due to crumpling the fabric when attempting to grasp 2 layers. Furthermore, the top layer of cloth can push downwards on the layer below it, which reduces the gap between the second and third layers; this reduced gap can make it difficult to grasp 2 layers. 
These observations and results suggest that further innovation on grasp policies may be necessary to improve 2-layer grasp performance on unseen cloths.

\section{Conclusion}\label{sec:conclusion}
% Daniel: I usually put a few future work directions here but Dave's guide suggests not to do this, so for now I just end on an optimistic note.

In this paper, we present a robotic system that uses magnetometer-based tactile sensing for precisely grasping layers of cloth. We train a classifier on tactile sensor readings from a ReSkin sensor. At test time, the classifier determines the number of layers of cloth grasped, which informs the policy to adjust the height of the gripper for subsequent grasp attempts. The system obtains strong results with grasping the top 1 or 2 cloth layers out of a stack of cloth, and generalizes to unseen cloth. We hope this work motivates future research on tactile-based robotic policies that can manipulate a wide variety of complex objects.

\section*{Acknowledgments}

% Daniel: trying to follow r-pad wiki guidelines.
{\footnotesize
\hl{
% We thank LG Electronics and the NSF CAREER grant number IIS-2046491 for funding. 
We thank LG Electronics, the NSF CAREER grant IIS-2046491, the NSF grant CMMI-2024794, and the NSF GRFP (DGE1745016, DGE2140739) for funding.
We thank our colleagues for helpful hardware support, in particular Sarvesh Patil, Raunaq Bhirangi, Tess Hellebrekers, and Pragna Mannam.  We thank our colleagues for helpful paper writing feedback, in particular Zixuan Huang, Chuer Pan, and Sarthak Shetty.}
}

% \clearpage Daniel: can do this if needed
%\footnotesize
% \bibliographystyle{IEEEtranS}
% \bibliography{example}
\renewcommand*{\bibfont}{\scriptsize}
\printbibliography

% Back to normal size.
\normalsize
\cleardoublepage
\appendices

%%%%%%%%%%%%
% APPENDIX %
%%%%%%%%%%%%

%\daniel{Wrapping the whole thing in the ``highlight'' for LaTeX was throwing errors -- but basically everything in the supplement is new so it might be overkill to explicitly highlight. Remove this and highlights for the final version (and arXiv)}

%\begin{itemize}
%    \item Appendix~\ref{app:method-details} provides more details on the methodology.
%    \item Appendix~\ref{app:additional-details} provides more details on the experiments.
%\end{itemize}

\section{ReSkin Data Details}\label{app:method-details}

In Section~\ref{ssec:classifier}, we describe the real-world procedure for collecting a training dataset of ReSkin sensor readings. 
Each training episode lasts for about 5 seconds to complete the approach, grasp, and release cycle. The grasp stage lasts for 1 second, resulting in approximately 350 sensor readings of 15 values each (3 per magnetometer). 
We filter out training episodes where no cloth was grasped.
Combining all episodes into a dataset results in a total of 18,838 ReSkin tactile sensor readings, each of which is 15-D.
The readings in the dataset are separated into training and validation sets, grouped by episode to prevent data leakage. Thus, all time steps in one episode are either all in training or all in validation.
We train classifiers taking individual readings as input and found that this input outperformed our baselines on the cloth singulation task; classification on time-series inputs are a promising area of future work.

\section{Image Classifier Details}\label{app:additional-details}

In Section~\ref{ssec:baselines}, we introduce the \textbf{Random-Image} and \textbf{Feedback-Image} baseline methods. These use images instead of tactile data, but the objective is the same: to predict the number of cloth layers grasped. 
When collecting tactile-based data, we simultaneously collect image data, so the data collection time for image data is the same as the time that was spent on collecting the Reskin data. We mount a webcam approximately \SI{30}{\centi\meter} away from the starting position of the ReSkin, which queries images of the robot when it is collecting data. See Fig.~\ref{fig:data} for examples of what the RGB images look like. 

The image classifier is an 18-layer ResNet~\cite{resnets_2016}. We use an off-the-shelf ResNet-18 from PyTorch which has been pre-trained on ImageNet, and change the last layer to output 4 dimensions instead of 1000. This results in a total of 11,178,564 trainable parameters.
The RGB input images are first cropped to $360\times 360$ such that the ReSkin sensor is located roughly in the middle, then images are further resized to $224\times 224$ before being passed as input to the ResNet.

There is about 7-8X more tactile data compared to image-based data because the ReSkin can query data much faster than the webcam. This may be an inherent advantage of the ReSkin sensor over most commercial webcams. To strengthen the image-based baseline, we use data augmentation (which the tactile classifiers do not use).
In training, we employ random crops by selecting random $224\times 224$ crops within the $360\times 360$ image. We also use random horizontal flips. At test time, we only use center crops for consistency. We train for 30 epochs, use a batch size of 64, and optimize using Adam with learning rate 0.001.

\end{document}